\documentclass[conference]{IEEEtran}
\usepackage{stfloats}
\usepackage{xcolor}
\usepackage{colortbl}
\usepackage[switch]{lineno}

\usepackage{url}
\usepackage{setspace}
\usepackage{placeins}
\usepackage{cite}

\usepackage{algorithm}
\usepackage[noend]{algpseudocode}
\algnewcommand{\LineComment}[1]{\State \emph{\textcolor{blue}{\(\triangleright\) #1}}}
\algrenewcommand\algorithmicindent{1em}%

\usepackage{graphicx}
\usepackage{caption}
\usepackage{subcaption}
\usepackage{tabularx}
\usepackage{enumitem}   

\usepackage{color, soul} 
\soulregister\emph7
\soulregister\cite7
\soulregister\ref7
\soulregister\pageref7

\definecolor{Gray}{gray}{0.85}
\definecolor{LightCyan}{rgb}{0.88,1,1}

\begin{document}

\newfloat{lstfloat}{htbp}{lop}
\floatname{lstfloat}{Listing}
\def\lstfloatautorefname{Listing} 

\setlength{\abovedisplayskip}{-1pt}
\setlength{\belowdisplayskip}{-1pt}

\setlength\floatsep{0.3\baselineskip plus 1pt minus 20pt}
\setlength\textfloatsep{0.3\baselineskip plus 1pt minus 20pt}
\setlength\intextsep{0.3\baselineskip plus 1pt minus 20pt}
\newcolumntype{R}{>{\centering\arraybackslash}m{3.5cm}}
\newcolumntype{L}{>{\centering\arraybackslash}m{1.5cm}}
\newcolumntype{M}{>{\centering\arraybackslash}m{3.5cm}}

\title{ML-SpecQD: Multi-Level Speculative Decoding with Quantized Drafts}

\author{\IEEEauthorblockN{Evangelos Georganas,
Dhiraj Kalamkar, Alexander Kozlov, 
Alexander Heinecke}
\IEEEauthorblockA{Intel Corporation}}

\maketitle

\begin{abstract}
 Speculative decoding (SD) has emerged as a method to accelerate LLM inference without sacrificing any accuracy over the 16-bit model inference. In a typical SD setup, the idea is to use a full-precision, small, fast model as “draft” to generate the next few tokens and use the “target” large model to verify the draft-generated tokens. The efficacy of this method heavily relies on the acceptance ratio of the draft-generated tokens and the relative token throughput of the draft versus the target model. Nevertheless, an efficient SD pipeline requires pre-training and aligning the draft model to the target model, making it impractical for LLM inference in a plug-and-play fashion. In this work, we propose using MXFP4 models as drafts in a plug-and-play fashion since the MXFP4 Weight-Only-Quantization (WOQ) merely direct-casts the BF16 target model weights to MXFP4. In practice, our plug-and-play solution gives speedups up to 2$\times$ over the BF16 baseline. Then we pursue an opportunity for further acceleration: the MXFP4 draft token generation itself can be accelerated via speculative decoding by using yet another smaller draft. We call our method ML-SpecQD: Multi-Level Speculative Decoding with Quantized Drafts since it recursively applies speculation for accelerating the draft-token generation. Combining Multi-Level Speculative Decoding with MXFP4 Quantized Drafts we outperform state-of-the-art speculative decoding, yielding speedups up to 2.72$\times$ over the BF16 baseline.
\end{abstract}

\section{Introduction}
\label{sec:introduction}
Large Language Models (LLMs) such as Chat-GPT~\cite{brown2020language} and GPT-4~\cite{achiam2023gpt}, have revolutionized the field of natural language processing (NLP) and have found application in areas like question-answering, coding, and user-interactive scenarios akin to chatbots and virtual assistants. As such, nowadays LLMs are being increasingly deployed on edge devices and PCs, giving birth to the era of ``AI PC" where the the models are deployed locally to assist the user in real time (e.g.\ as standalone AI assistants in coding~\cite{chen2021evaluating}) or as part of more complicated software/workload workflows (e.g.\ LLMs used internally in other application for documents and image processing~\cite{nguyen2024instruction}). In all these cases, in order to improve user-experience the LLMs shall not sacrifice accuracy, and at the same time the latency should still stay within reasonable limits for real-time interaction.

Unfortunately, the rising size of LLMs implies ever-growing latency due to their auto-regressive nature: they generate tokens one by one. A recent solution to mitigate this issue is called speculative decoding (SD)~\cite{leviathan2023fast,chen2023accelerating,xia2022speculative}. In a typical SD setup, the idea is to use a small, fast model as “draft” to generate the next few tokens and use the “target” large model to verify the draft-generated tokens. The verification step comprises merely of a single forward pass of the target model. After verification, the draft-generated tokens following the first mis-matched token are discarded and the process continues with the next batch of draft-tokens to be generated. By leveraging this methodology, one can leverage a smaller, much faster model to generate many of the desired output tokens whereas the output is virtually identical to the one from large “target” model. As a result, SD does not compromise accuracy and still it enjoys substantial latency reduction over the traditional greedy decoding strategy.

\begin{figure}[t]
\centering
\includegraphics[width=\columnwidth]{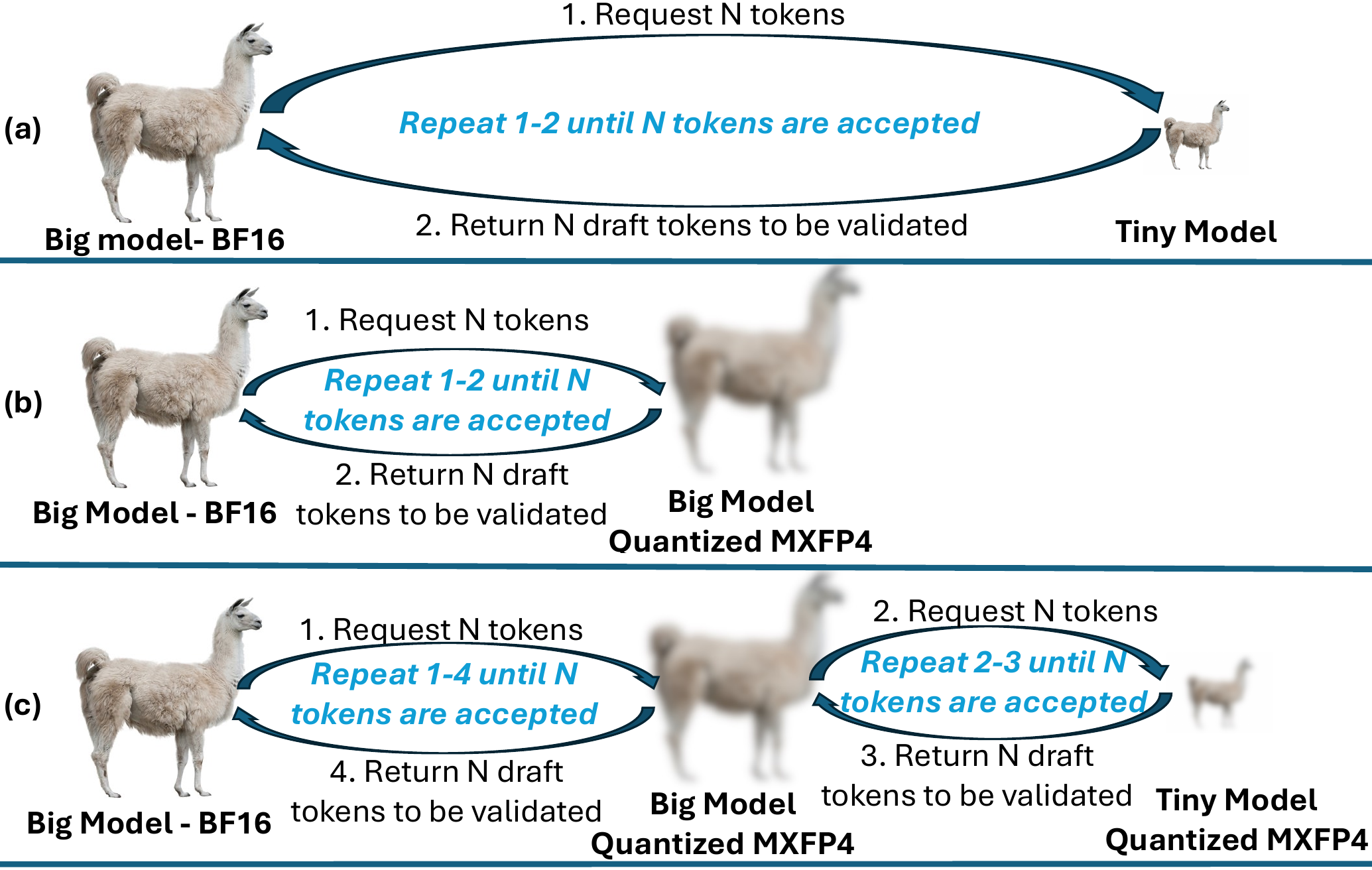}
\caption{Various flavors of Speculative Decoding: (a) Speculative decoding with a large target model (e.g.\ Llama 7B in BF16) and a custom, small draft model (e.g.\ Tiny Llama 68M), (b) Speculative decoding with a large target model and an MXFP4-direct-cast-quantized target model as draft model, (c) Multi-Level Speculative Decoding: a large target model uses an MXFP4-direct-cast-quantized target model as draft model, and subsequently the MXFP4-direct-cast-quantized model uses a smaller draft, potentially also MXFP4-quantized.}
\label{fig:ml_spec}
\end{figure}

The common practice in  SD is to use a very small draft model, around two orders of magnitude smaller than the target model, in order to observe substantial speedups~\cite{leviathan2023fast}, since the draft model itself suffers from the same auto-regressive nature of generation. The latency issue for the drafting model is de-emphasized if it is noticeably smaller than the target model. Nevertheless, this common practice exhibit two drawbacks. First, a small draft is not always readily available; for example, the smallest model in the Llama2 foundational family has 7 billion parameters~\cite{touvron2023llama}. The community has therefore developed methods (e.g.\ FastDraft~\cite{zafrir2024fastdraft}) to obtain such small draft models, but these techniques require pre-training and aligning the draft model to the target model, making them impractical for LLM inference in a plug-and-play fashion. Second, even in case such a small model is readily available, usually its direct usage as draft results in poor acceptance of the draft-generated tokens by the target models given the capacity gap between the target and draft~\cite{miao2024specinfer}. As a consequence of the poor acceptance ratio of the draft-generated tokens, the larger and much slower target model has to be called an increasing number of times for verification/validation, and eventually the performance benefits of SD are diminished.

We make two key observations with respect to the limitations of SD as described in the previous paragraph. First, recent advances in Weight-Only-Quantization (WOQ) with 4-bit datatypes (and beyond) has shown to generate high quality results with small accuracy loss over the BF16 models~\cite{dettmers2023qlora,rouhani2023microscaling,lee2024amxfp4,lo2024nanoscaling}. Typically one can leverage such WOQ methods without any fine-tuning/extra processing overhead by merely direct-casting the weights to the target precision, making them a plug-and-play solution for inference acceleration especially in bandwidth-bounds scenarios~\cite{shen2023efficient}. Therefore, we propose the usage of 4-bit quantized models as drafts in a SD setup, since they illustrate substantial speedups over the target BF16 models, and additionally they can yield high acceptance ratios of the draft-generated tokens given their intrinsic alignment with the base/target model (see Figure~\ref{fig:ml_spec}(b)). The second key observation is inspired by multi-grid methods in HPC~\cite{hackbusch2013multi} where one considers a problem on different refinement levels and uses solutions on coarser levels to improve upon solutions on finer levels. In the context of inference via conventional SD, a custom, tiny model is available as draft (i.e.\ ``coarse" model), and we use it directly from the large target model (i.e.\ ``fine" model) for token-drafting in one-step, and inadvertently suffer from the low acceptance ratios of the coarsely-drafted-tokens (see Figure~\ref{fig:ml_spec}(a)). In this work, we propose the introduction of an intermediate model for token-drafting between the target and the tiny models that is merely a 4-bit direct-cast of the target model (thus its granularity is between the ``fine" target model and the ``coarse" draft model). In such a setup, the target model uses a 4-bit direct-cast-model as draft, and subsequently the intermediate 4-bit model uses the tiny model for its own speculative decoding (i.e.\ we apply multi-level, hierarchical decoding illustrated in Figure~\ref{fig:ml_spec}(c)). The benefit of this multi-level approach is two-fold: First, it mitigates the issue of low acceptance ratios of the tiny-draft-tokens that result in excessive number of invocations of the target BF16 model, since now their verification is subsumed by the intermediate, much faster 4-bit draft model. Second, the intermediate model's token-drafting itself is accelerated by the tiny draft, making the multi-level SD a plug-and-play solution in single-draft SD setups where a tiny draft is available.

We call our proposed solution ML-SpecQD: Multi-Level Speculative Decoding with Quantized Drafts and we illustrate its overview in Figure~\ref{fig:ml_spec}. Essentially ML-SpecQD combines the ideas of multi-level decoding and quantized drafts. Unlike recently proposed multi-stage methods that require \emph{multiple} custom models (one model for each level of drafting)~\cite{spector2023accelerating,chen2023cascade}, ML-SpecQD reaps the benefits of multi-level decoding by merely having a single tiny draft, and leverages direct-cast-quantized models as intermediate drafts. Also, unlike recent work that uses a \emph{single} level of self-speculation with an int4 quantized draft~\cite{tiwari2025quantspec}, ML-SpecQD can further accelerate the entire decoding pipeline by accelerating the 4-bit drafting with a much smaller draft-model in case one is available. Even though we present ML-SpecQD with 2 levels of hierarchical drafting, it is naturally extended to more levels where the drafting at each level can be a quantized model of the previous levels or any kind of draft that is used in traditional SD (e.g.\ a small statistical model, self-speculating models). In this work we focus on MXFP4 models as intermediate draft and we optimize the computational kernels on CPUs such that ML-SpecQD can be readily employed on AI PC devices. Nevertheless, the ML-SpecQD framework is applicable with any low-bit quantization method (e.g.\ 8-bit, 4-bit, 1.58 bit, 1 or 2-bit) and any acceleration platform (e.g.\ GPU/NPU). For example, one could use as target model a foundational BF16 LLM, as level-1 draft model a 4-bit quantization of the target model, and as level-2 draft model a 1-bit quantization of the target model. All these drafts are intrinsically aligned with the target model (conforming to the quantization scaling principles of LLMs~\cite{kaushal2024spectra}), and recent quantization advances exhibit solid tradeoffs between quantization bit-widths and sustained accuracy~\cite{liu2025paretoq}. The contributions of this work are:
\begin{itemize}
\item The introduction of 4-bit LLM quantization (leveraging MXFP4 datatype) as a viable, hardware-agnostic, readily available, plug-and-play solution for the draft model in SD pipelines, yielding high draft-token acceptance ratios and accelerating the end-to-end LLM inference.
\item The proposal of a multi-level speculative decoding framework namely ML-SpecQD, where low-bit quantized models play a pivotal role as drafts. This hardware-agnostic framework combines the performance benefits of hierarchical SD and LLM quantization \emph{without} the need for \emph{multiple}, custom draft models that require costly pre-training and alignment with the target model.
\item A high-performance MXFP4 GEMM microkernel that is the main computational engine for the tensor contractions in the MXFP4 quantized drafts of ML-SpecQD for client CPU/edge inference platforms.
\item An experimental analysis on client CPU of speculative decoding with quantized drafts and multi-level speculative decoding with ML-SpecQD, exhibiting LLM inference speedups up to 2.72$\times$ over the BF16 baseline, without sacrificing any accuracy/yielding same output as BF16 LLM inference.
\end{itemize}

\section{Related Work}
\label{sec:related}
\subsection{Accelerating LLM Inference}
In the past few years, several methods have been developed by academia and industry in order to reduce LLM inference costs~\cite{treviso2023efficient}, minimize latency and increase computational efficiency, including sparsification/pruning~\cite{frantar2023sparsegpt}, knowledge distillation~\cite{hinton2015distilling,taori2023stanford} and quantization~\cite{lin2024awq, chee2023quip, liu2025paretoq}. Sparsification effectively compresses the models by setting tensor entries to zero in a structured or unstructured fashion, and as a result the compressed model can save storage space and under certain scenarios it may accelerate even the LLM inference pipeline (e.g.\ when sparsity is supported by hardware or in bandwidth-bound scenarios). Orthogonal to sparsification, knowledge distillation is transferring knowledge from a larger model to a smaller model and this process can result in small models that are quite accurate for certain domain-specific tasks (e.g.\ such small models can be used as drafts in SD~\cite{zafrir2024fastdraft}). Finally, quantization is reducing the precision of model weights (e.g.\ from 16-bits to 4-bits) to reduce the memory requirements of the model, and also potentially accelerate the inference in cases the underlying hardware provides specialized instructions or in case the pipeline operates in a bandwidth-bound regime. However, quantization comes at a cost of reduced accuracy~\cite{rouhani2023microscaling} which might not be acceptable for some use-cases.

\subsection{Speculative decoding}
A recent solution to accelerate LLM inference is called speculative decoding (SD), and does so by using a small, fast model as “draft” to generate the next few tokens and use the ``target" large model to verify the draft-generated tokens~\cite{leviathan2023fast, chen2023accelerating, xia2022speculative ,xia2024unlocking}. In such a setup, the output of the LLM inference is tracking the output of the ``target" model, thus there is no impact on accuracy (one of the main drawbacks of LLM model quantization), while the end-to-end latency is significantly reduced. One extension of SD is enhancing the model to produce multiple tokens at once~\cite{cai2024medusa, li2024eagle, bhendawade2024speculative}; such methods are not plug-and-play since they require fine-tuning.

Self-speculative decoding is a methodology that restricts the draft model to be of the same architecture like the target and shows benefits due to the intrinsic alignment of the models~\cite{sun2024triforce, xia2024unlocking}. Recent work has also proposed int4 quantization of the model and a hierarchical quantization of the KV-cache within a SD setup~\cite{tiwari2025quantspec}, however it focuses on a single level of speculation. Also, recent work has explored the acceleration of int4 weight-only quantized targets that use high-precision activations by using 4-bit joint weight-activation quantization schemes as draft models~\cite{zhao2025qspecspeculativedecodingcomplementary}. However, this setup does not yield any speedup in memory bound use-cases where the model reading is the bottleneck (e.g.\ low-batch inference) since both the target and the draft use the same set of weights. The same prior work also focuses on a single level of speculation, and the starting point is a 4-bit model which might yield substantial accuracy drop on specific tasks.

Another recent extension of SD relies on leveraging \emph{multiple} custom drafts of different sizes in a hierarchical fashion to further reduce the LLM inference latency~\cite{spector2023accelerating,chen2023cascade}. Unlike these hierarchical methods that require an impractical number of smaller, custom drafts, our work reaps the benefits of multi-level decoding by merely having a single tiny draft, and leverages direct-cast-quantized models as intermediate drafts that are readily available via direct-casting of the target model. In other words, our work uniquely brings plug-and-play quantization methods within multi-level SD setups to maximize performance without the requirement of pre-training/aligning multiple draft models.

\section{The ML-SpecQD Framework}
\subsection{Speculative Decoding Overview}
In a typical speculative decoding setup, the idea is to use a small, fast model as “draft” to generate the next $N$ tokens (Figure~\ref{fig:ml_spec}(a)-arrow 1), and then use the “target” large model to verify the $N$ draft-generated tokens (Figure~\ref{fig:ml_spec}(a)-arrow 2). The verification step comprises merely of a single forward pass of the target model. After verification, the draft-generated tokens following the first mis-matched token are discarded and the process continues with the next batch of draft-tokens to be generated. By leveraging this methodology, one can leverage a smaller, much faster model to generate many of the desired output tokens whereas the output is virtually identical to the one from large “target” model. Therefore, SD does not compromise accuracy and still it enjoys substantial latency reduction over the traditional greedy decoding strategy.

\label{sec:framework}
\subsection{Rationale for Quantized Drafts}
\label{subsec:rationale}
\begin{figure}[t]
\centering
\includegraphics[width= \columnwidth]{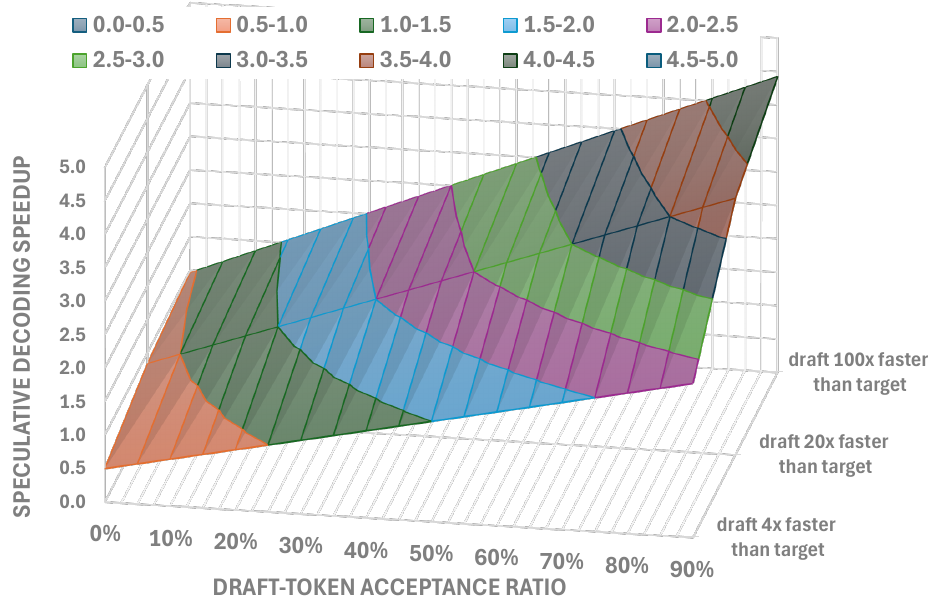}
\caption{Expected speculative decoding speedup for different draft-token acceptance ratios, for 3 different drafts that are $4\times$, $20\times$ and $100\times$ faster (smaller) than the target model.}
\label{fig:rationale}
\end{figure}
For the presentation of the ML-SpecQD framework and the reasoning of its performance we will make two assumptions that are \emph{not} inherent to the ML-SpecQD methodology. First, we will focus on the analysis of a single user/single batch LLM inference that is common in personalized applications and minimize the latency of the workflow (e.g.\ AI PC applications). Nevertheless, extending SD with larger batch-size support has been explored in recent work and is an active research area~\cite{su2023synergy, liu2024optimizing}. Second, we focus on models with relatively shorter context length (e.g.\ up to 4k). Under these two assumptions (single batch and relatively shorter context length) the decoding phase is dominated by the linear projections in the model and exhibits low arithmetic intensity, making the performance of the pipeline typically bandwidth bound and dominated by the GEMMs in linear layers~\cite{tiwari2025quantspec}. In case of larger context length where the attention starts dominating the overall performance, we can leverage the techniques of recent work that quantize the KV-Cache and accelerate the decoding phase~\cite{tiwari2025quantspec,li2024snapkv,chen2024magicdec}. Therefore, techniques accelerating SD in scenarios with larger batch size and longer context can be seen as complementary to our ML-SpecQD framework.

Under the assumptions mentioned in the previous paragraph, we can simply correlate the decoding performance of an LLM model with its model size: the performance is dominated by the linear-projections that exhibit low arithmetic intensity, making the decoding phase bandwidth-bound by essentially ``streaming in" the model weights. For the sake of simplicity let's assume that a model $S\times$ smaller than a target model will also run decoding $\approx S\times$ faster. As such, we can expect that a 4-bit quantized model of a target 16-bit model is going to be almost $4\times$ faster than the base models since it is by construction $\approx 4\times$ smaller. Similarly, a tiny draft model that is $100\times$ smaller than the target model is expected to be $\approx 100\times$ faster. However, the efficacy of a SD pipeline does not merely depend on the relative speedup of the draft over the target model. A key metric that affects the overall SD efficiency is the \emph{acceptance ratio} of the draft-generated tokens. Using the example of Figure~\ref{fig:ml_spec}(a), out of the $N$ draft-generated tokens, the target model may accept only the first $G$ tokens, and effectively the acceptance ratio for this round of SD is $\alpha=G/N$. For this specific round, the SD pipeline produced $G+1$ tokens (since the forward validation step produces an extra token) and it required 1 execution of the target model and $N$ executions of the draft model. Assuming that the draft model is $S\times$ faster than the base model, the expected speedup of SD over greedy decoding for this round is:

\begin{equation}
Speedup_{SD} = \frac{G+1}{1+N/S}=\frac{\alpha+1/N}{1/N+1/S}
\label{eqn:speedup}
\end{equation}

By setting e.g. $N=4$ in the previous equation, we get the expected speedup to be $(\alpha+0.25)/(1/S+0.25))$. In Figure~\ref{fig:rationale} we plot the expected SD speedup for different draft-token acceptance ratios and for 3 different drafts that are $4\times$, $20\times$ and $100\times$ faster (smaller) than the target model. Due to the non-linear nature of Equation~\ref{eqn:speedup}, we make an interesting observation: a draft model that is $4\times$ faster than the target and results in high draft-token acceptance ratio $\alpha$ (e.g.\ 80\%-90\%) can yield in SD setup similar speedups to those achieved by much smaller draft models (e.g $20\times$ and $100\times$ smaller than target) with draft-token acceptance ratios $\approx \alpha/2$ (i.e. see magenta band in the surface plot of Figure~\ref{fig:rationale}). This observation implies that 4-bit quantized models of the 16-bit target model that are readily available via weight-only-quantization, and maintain high accuracy~\cite{dettmers2023qlora,rouhani2023microscaling,lee2024amxfp4,lo2024nanoscaling}, can serve as plug-and-play drafts in a SD setup (see Figure~\ref{fig:ml_spec}(b)), obviating the need for tiny, custom drafts which fundamentally yield much lower acceptance ratios (due to capacity and alignment issues with the target model). Here we mention the 4-bit quantization as an illustrative example; recent advances in quantization has shown that proper 1, 1.58, 2, and 3 bit quantization schemes can yield models with good accuracy~\cite{liu2025paretoq}, and for our purposes these quantized models could be used as drafts within SD pipelines with even bigger performance upside.

\subsection{Multi-Level Decoding Architecture}
\label{subsec:architecture}
\begin{figure}[t]
\centering
\includegraphics[width= \columnwidth]{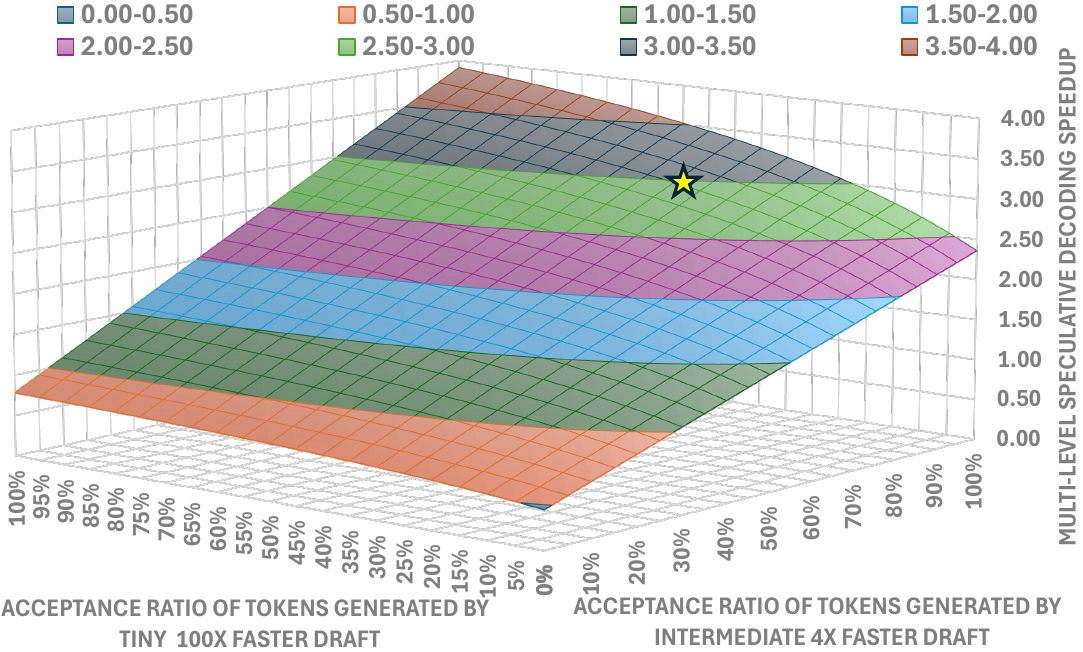}
\caption{Expected speedup in a 2-level speculative decoding setup, with a large BF16 target model, a 4-bit quantized model (4$\times$ faster than target) as intermediate draft, and a small model (100$\times$ faster than target) as a last-level draft.}
\label{fig:multispec}
\end{figure}
Now that we have a plug-and-play, readily-available draft model that is 4$\times$ smaller than the target, a natural question is: \emph{Can we do better in case that yet another smaller, tiny draft model is available?} To take advantage of both the 4$\times$ smaller draft and the even smaller draft model, we propose a multi-level architecture. More specifically, we propose the introduction of the 4-bit direct-cast of the target model (and thus 4$\times$ smaller) as an intermediate model for token-drafting between the target and the tiny model. In such a setup, the target model uses a 4-bit direct-cast-model as draft, and subsequently the intermediate 4-bit model uses the tiny model for its own speculative decoding (i.e.\ we apply multi-level, hierarchical decoding illustrated in Figure~\ref{fig:ml_spec}(c)). The benefit of this multi-level approach is two-fold: First, it mitigates the issue of low acceptance ratios of the tiny-draft-tokens that result in excessive number of invocations of the target BF16 model since now their verification is subsumed by the intermediate, much faster 4-bit draft. Second, the intermediate model's token-drafting itself is accelerated by the tiny draft model. In order to get an estimate of the upside in such a hierarchical setup, we can again use Equation~\ref{eqn:speedup}. However, now the term $S$ in the denominator is not simply $S=4$, instead it is a value $S' > S$ since the intermediate 4-bit model is also accelerated by SD with the smaller, tiny draft. In Figure~\ref{fig:multispec} we illustrate the expected speedups in such a multi-level decoding architecture. More specifically, we illustrate the performance upside in a 2-level SD setup, with a large BF16 target model, a 4-bit quantized model (4$\times$ faster than target) as intermediate draft, and a small model (100$\times$ faster than target) as a last-level draft. The x/y axes show the acceptance ratios of the tokens generated by the last level draft (left) and the acceptance ratios of the tokens generated by the 4-bit quantized draft (right). This surface plot illustrates the compound effect of multi-level drafting: For acceptance ratios in the regime of 80-85\% for the 4-bit quantized draft and acceptance ratios of 40-45\% for the tiny-draft generated tokens, we expect speedups in the range 3-3.5$\times$ over the target model (see star-annotated data point in plot). If we were to use just 1-level decoding with either the 4-bit draft or the tiny-model with the same acceptance ratios, we would get speedups in the range of 2-2.5$\times$ (see magenta band in Figure~\ref{fig:rationale}).

Even though for presentation purposes we focused on a 2-level setup, ML-SpecQD generalizes to multiple levels of decoding where each level can use \emph{any} kind of draft that may accelerate the decoding of the previous level. For instance it can be a quantized draft model of the previous level (either obtained via direct-cast or via fine-tuning methods~\cite{liu2025paretoq}), a smaller custom fine-tuned model (e.g.\ obtained via FastDraft~\cite{zafrir2024fastdraft}), models relying on self-speculation~\cite{zhang2023draft} or even simple statistical language models~\cite{chen2023cascade}.

\subsection{Implementation Details}
\label{subsec:implementation}
\subsubsection{Assistant Decoding in Hugging Face Transformers}
\label{subsubsec:assistant}
We implemented ML-SpecQD on top of the Assisted Decoding framework in HuggingFace (HF) Transformers~\cite{wolf2020transformers}. More specifically, we extend the infrastructure that currently accepts only one draft model (and as such simple one-level speculative decoding) to be able to get multiple drafts. Each draft may have its own tokenizer by embracing the Universal Assisted Generation path in Transformers. Therefore, the API changes to existing HF speculative decoding, are minimal (i.e.\ we just provide multiple models in the \emph{generate} API function and internally it triggers multi-level decoding). We note that based on the configuration provided in the \emph{generate} API call, one may leverage self-speculation methods already supported in HF framework as part of the multi-level speculation (e.g.\ early-exit inference~\cite{elhoushi2024layerskip} and prompt-lookup decoding~\cite{saxena2023prompt}).

\subsubsection{PyTorch Extensions for LLM inference}
\label{subsubsec:tpp_ext}
In order to maximize the performance within HF Transformers, we use the Tensor Processing Primitives (TPP) PyTorch C++ extensions backend~\cite{georganas2021tensor, georganas2024harnessing}. The TPP PyTorch C++ extensions invoke tensor contractions over blocked tensor layouts leveraging the BRGEMM TPP~\cite{georganas2020harnessing} which in turn uses the libxsmm~\cite{heinecke2016libxsmm} backend for Just-In-Time (JIT) GEMM code generation. Additionally, the TPP PyTorch extension fuses bias, dropout, residual add and layernorm on a small 2D-block granularity to maximize the out-of-cache-reuse of tensors among subsequent operators. Last but not least, the latest release of TPP PyTorch C++ extensions internally uses the PARLOOPER framework~\cite{georganas2024harnessing} for JIT nested-loop code generation, allowing optimal parallel execution on hybrid platforms with performance and efficiency cores. Previous work has shown that such a setup in HF outperforms the default oneDNN~\cite{georganas2024harnessing} backend on CPUs for LLM inference by up to 1.3$\times$. In the next subsection we detail the new MXFP4 GEMM microkernel we designed and integrated in the TPP PyTorch C++ extension in order to utilize it within ML-SpecQD whenever MXFP4 quantized drafts are deployed.

\subsubsection{MXFP4 direct-cast of weights and GEMM Microkernel}
\label{subsubsec:mxfp4_gemm}
\begin{figure}[t]
\centering
\includegraphics[width=1.1\columnwidth]{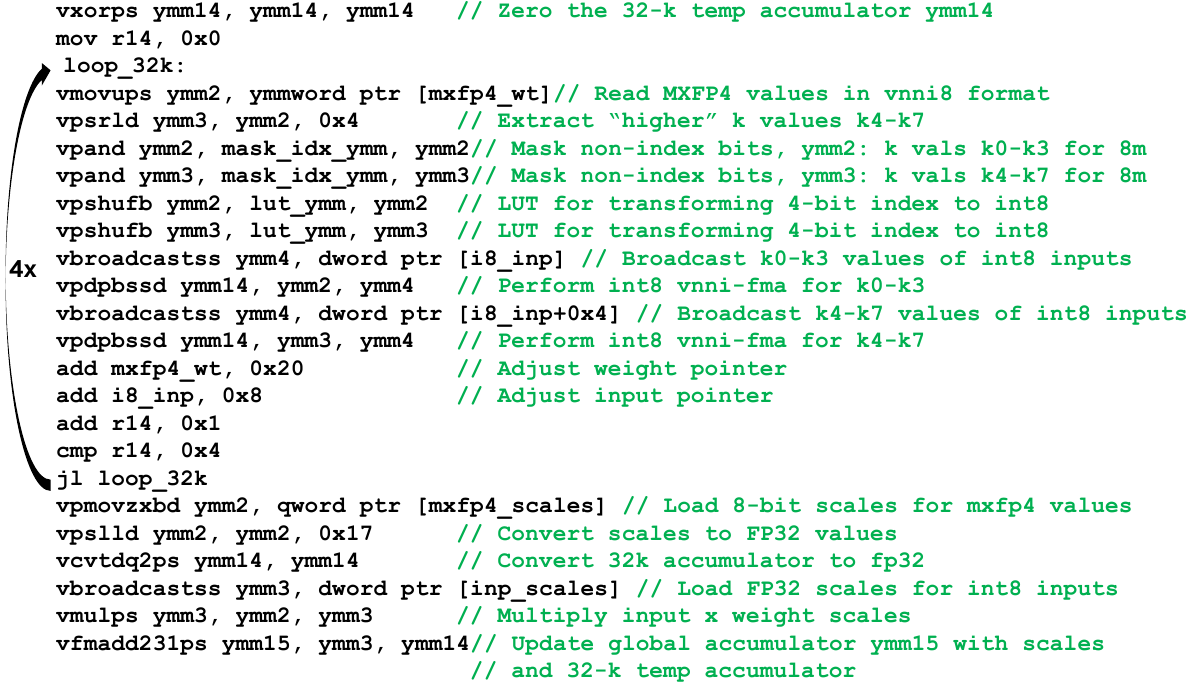}
\caption{AVX2 GEMM microkernel with MXFP4 weights and vnni-INT8 FMAs ($M=8$, $N=1$, $K=32$).}
\label{fig:microkernel}
\end{figure}

In this section we detail the heart of the computation in our MXFP4 quantized token-drafting, namely the MXFP4 GEMM microkernels. Specifically we focus on modern CPUs with AVX2 instructions that can be found on contemporary AI PC devices. As described in Section~\ref{subsec:rationale}, we are focusing on the use-cases where the GEMMs illustrate algorithmically very low arithmetic intensity, thus are typically bandwidth bound. Also, we focus on the MXFP4 \emph{direct-cast} weight quantization that has shown good accuracy in LLM inference and does not require any special quantization recipes or fine-tuning, making them a truly plug-and-play solution.

Per specification of the MXFP4 standard~\cite{ocp}, we quantize offline the model weights from the source precision, i.e. BF16, down to MXFP4 in two steps. For each 32-element vector $V$ of BF16 weight values: i) We set $X$ to be the largest power-of-two less than or equal to the maximum absolute value of $V$, divided by the largest power-of-two representable in the element data type, ii) For each element $v_i$ in $V$ we perform the scaling $v_i/X$ quantized to the target FP4 datatype E2M1 (meaning we use 1 bit for sign, 2 bits for exponent and 1 bit for mantissa). For this quantization step, normal numbers that exceed the max normal representation of the element datatype are clamped to the max normal, preserving the sign. This direct-cast quantization happens offline during model loading/preparation and does not need to be performant since it is not on the critical path during inference. The outcome of this 2-step direct-cast quantization is a set of an FP4 weight tensor along with an 8-bit scale factor tensor. We get one 8-bit scale factor per 32 FP4 elements, where the 8-bit scale factor follows the format E8M0, i.e.\ it is an unsigned representation of a conventional biased 32-bit float exponent.

During inference, we have to read from memory these two tensors of FP4 values and scales, effectively reducing the BW requirements by a factor of 3.76$\times$ compared to the case with BF16 weights. However, reading these MXFP4 weights from memory, performing the required multiplications/additions (FMAs) and hitting the bandwidth limits is challenging for two reasons. First, up-converting naively the FP4 weights to FP32 values (e.g.\ via look-up tables/permutes), multiplying them with the corresponding scales (after up-converting them to FP32) to get FP32 weights values, and finally multiplying the up-converted FP32 weight values with the input activations requires an excessive number of operations that would diminish any performance benefits from reading the low-bit data type. In other words, we increase the operational intensity since we \emph{repeatedly} have to multiply the scales with the FP4 values. Second, the CPUs on modern AI PC platforms have a limited number of cores (e.g.\ 10-20s), yet the available bandwidth is in the order of 100s of GB/s. Therefore, the available bandwidth per core is quite high, meaning that simple FP32 FMA throughput is not sufficient to fully-saturate the available read-memory bandwidth per core (especially as we read $\approx$3.76$\times$ less data than the BF16 equivalent GEMMs), again negating any benefits from reading the lower-bit datatype.

In order to mitigate the two aforementioned issues, we propose the microkernel in Figure~\ref{fig:microkernel}. First, we follow, a late-scaling algorithm where we apply the MXFP4 micro-scaling on the partial accumulators rather than the input weight vectors (i.e.\ after the partial 32-length accumulation has happened). This formulation is mathematically equivalent and does not compromise any accuracy, yet it saves redundant/repeated FP32 multiplications with the scales. For multiplying and adding (a la inner-product fashion) 32 weight values $w_i$ having the same shared micro-scale factor $scf$ with 32 activation values $a_i$ we have the following equation for the partial accumulator:
\begin{equation}
acc_{partial} = \sum_{i=0}^{31}scf\cdot w_i \cdot a_i = scf\cdot \sum_{i=0}^{31}w_i \cdot a_i
\label{eqn:late_acc}
\end{equation}

Second, in order to take advantage of the modern int8 VNNI FMAs on modern AVX2 CPUs, we opt to further quantize the conversion lookup-table FP4$\rightarrow$F32 to int8 values, i.e.\ the lookup-table now maps FP4$\rightarrow$int8 values. Subsequently, by quantizing the input activations to int8 as well we can readily employ the int8 VNNI FMAs and ameliorate the FMA pressure of the GEMM microkernel (the int8 VNNI FMAs exhibit 4$\times$ the throughout of the FP32 FMAs).

In short, the GEMM microkernel illustrated in Figure~\ref{fig:microkernel} first loads the 4-bit weight values stored in vnni-8 format, performs lookups using byte-permutes to convert the 4-bit values (that are essentially the indices of the permute) to int8 values, and then performs vnni-int8 FMAs with the activations. Once 32 weight values $\times$ activation values have been accumulated in the inner product (see equation~\ref{eqn:late_acc}), we read the weight scaling factors along with the input activation scale factors, we scale the partial accumulators, and update the running/global accumulator vector registers. As we show later in Section~\ref{subsec:mxfp4_perf}, this microkernel is essential to maximize the MXFP4 GEMM performance during LLM inference. We implemented this microkernel in the libxsmm backend. Last but not least, our multi-threaded GEMMs within the PyTorch TPP extensions use this MXFP4 GEMM microkernel as building block, and for the parallelization portion we leverage PARLOOPER and its hybrid scheduling capabilities to fully utilize all performance and efficiency cores~\cite{georganas2024harnessing}.

\subsection{Limitations}
\label{subsec:limitations}
In this subsection we would like to mention a few limitations of the ML-SpecQD framework. First, even though the memory overhead of the MXFP4 quantized draft is generally negligible ($\approx 25\%$ assuming a BF16 target model), under certain scenarios with tight memory constraints this extra overhead might be the limiting factor. Second, the performance benefits presented in this work are expected to be maximized in cases where the inference is limited by memory bandwidth as explained in Section~\ref{subsec:rationale}. The crux of speculative decoding is that the verification step of multiple draft tokens comes at the same cost of a single forward pass of the greedy decoding. This assumption is not valid for cases where the draft-verification is not memory-bound. Third, the benefits of ML-SpecQD are expected to be larger when the draft-token acceptance ratios of the MXFP4 draft and the small draft differ by a large margin (see Figures~\ref{fig:rationale} and~\ref{fig:multispec} for the the amplified speedups when the acceptance ratios differ substantially). For example, if there is available a small model (e.g.\ 10$\times$ smaller) yielding high acceptance ratios for the use-case case of interest (e.g.\ 70\%), the benefits of MXFP4 quantized drafts and ML-SpecQD might be marginal over traditional speculative decoding.

\section{Results}
\label{sec:results}
\subsection{Experimental Setup and Platform}
We assess the performance of ML-SpecQD on an Intel Core Ultra 9 285K CPU with 24 cores (8 performance and 16 efficiency cores) and the max turbo frequency is 5.2 Ghz. On this platform we have an overclocked memory DDR5@7200 MT/s with read bandwidth measured at 104 GB/s. For the evaluation we designed two experimental setups:
\begin{itemize}
\item A Question-Answering (QA) setup using as target model the BF16 Llama2 7B model which is finetuned for chat~\cite{touvron2023llama}. This is the smallest model in the foundational suite of Llama2 models, therefore as a custom, small draft model we use a 68M parameter Llama model fine-tuned for chat scenarios. This small, Llama-like model has been previously developed and used for drafting in SD~\cite{miao2024specinfer}. It is trained on Wikipedia and part of the C4-en and C4-realnewslike datasets. For the performance evaluation we used 100 prompts from the HAGRID dataset~\cite{kamalloo2023hagrid}. 
\item A code generation benchmark using as target model the BF16 Qwen2.5-Coder 7B which is a state-of-the-art open-source code model~\cite{hui2024qwen2}. The Qwen2.5-Coder family of models comes  with a 0.5B parameter model which we will be using as small draft in our SD assessment. For the performance evaluation we used 100 prompts from the MBPP dataset~\cite{austin2021program}.
\end{itemize}
For all the evaluations with SD, we setup the generation configuration to use greedy search with 8 assistant draft tokens (speculation length). For the cases in ML-SpecQD where the draft is not sharing the same tokenizer with the previous-level model (i.e.\ the use-case with the Qwen2.5 models) we use a more stringent confidence threshold of 0.65 compared to the default value of 0.4. Tuning confidence thresholds/parameters for the acceptance of draft tokens, or deploying recent methods that dynamically adjust the speculation length~\cite{mamou2024accelerating} could even further speedup the inference pipeline. An exhaustive co-tuning of such generation configuration parameters is beyond the scope of this paper and is subject of future work.

\subsection{Evaluation on Question-Answering with Llama2 7B}
\begin{figure*}[t!]
\centering
\includegraphics[width=2.15\columnwidth]{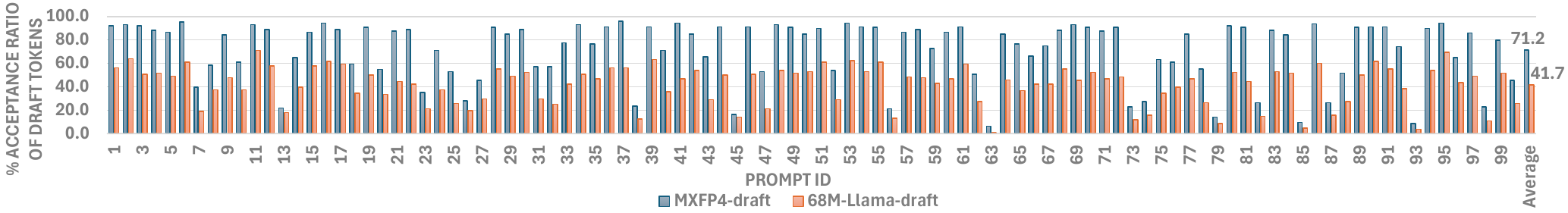}
\caption{QA benchmark: Acceptance ratios (\%) of draft tokens in two speculative decoding setups: (a) MXFP4-direct-cast-quantized draft (blue bars) and (b) 68M custom Llama draft (orange bars) over 100 input prompts.}
\label{fig:accept_llama}
\end{figure*}

\begin{table}[t]
  \begin{center}
    \begin{tabular}{l| c}
      \textbf{Inference Pipeline} & \textbf{Geomean Speedup}\\\hline \hline
      i) Greedy Decoding (baseline) & 1.00$\times$\\ \hline
     ii) SD  with 68M draft & 1.81$\times$\\ \hline
    iii) SD with MXFP4 draft & 1.86$\times$\\ \hline
    iv) Multi-level SD & \textbf{2.22$\times$}\\ \hline
    \end{tabular}
  \end{center}
\caption{Speculative decoding speedups on the QA use-case with Llama2 7B BF16 model as baseline. We show with bold text the best performing strategy.}
\label{tab:llama}
\end{table}

For the evaluation on Question-Answering we benchmarked the following generation strategies: (i) BF16 Greedy decoding with Llama 7B (i.e.\ our baseline), (ii) Speculative decoding using the 68M model as draft, (iii) Speculative decoding using an MXFP4-direct-cast model of the 7B target model as draft, and (iv) Multi-level speculative decoding with the first-level draft being an MXFP4-direct-cast model of the 7B target model and the 68M model as second-level draft.

Table~\ref{tab:llama} illustrates a summary of the experimental results. First we observe that that SD with the 68M custom draft yields a geomean speedup of 1.81$\times$ over the greedy decoding of the 7B models. The SD with the MXFP4-direct-cast model of the 7B target model as draft yields comparable speedup of 1.86$\times$. It is worth noting though that the later is a readily available model via direct casting, unlike the custom 68M model that requires pre-training. In order to further shed light on these result, we illustrate in Figure~\ref{fig:accept_llama} the draft acceptance ratios for these two SD setups over the 100 input prompts (strategy (ii) is orange bars and strategy (iii) corresponds to blue bars). We observe that the method with the MXFP4-direct-cast-quantized draft illustrates on average 1.7$\times$ higher acceptance ratio than the method with the custom/small 68M model draft (see 71.2\% vs 41.7\%). These quantitative results are validating the SD speedup projections illustrated in Figure~\ref{fig:rationale}: a draft model that is $4\times$ faster than the target and results in high draft-token acceptance ratio $\alpha$ can yield in SD similar speedups to those achieved by much smaller draft models (e.g\ $20\times$ -- $100\times$ smaller) with draft-token acceptance ratios $\approx \alpha/2$. Finally, we observe that using multi-level decoding (strategy (iv)) can further speedup method (iii) by using the Llama 68M model as a second-level draft, yielding eventually speedup of 2.22$\times$ over the baseline greedy decoding.

\subsection{Evaluation on Code Generation with Qwen2.5-Coder 7B}
\label{subsec:code_exp}
\begin{figure*}[t!]
\centering
\includegraphics[width=2.15\columnwidth]{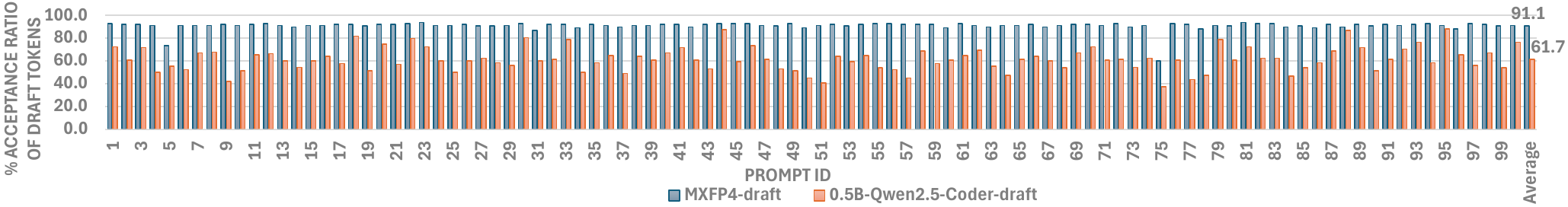}
\caption{Code generation benchmark: Acceptance ratios (\%) of draft tokens in two speculative decoding setups: (a) MXFP4-direct-cast-quantized draft (blue bars) and (b) Qwen2.5-Coder 0.5B draft (orange bars) over 100 input prompts.}
\label{fig:accept_qwen}
\end{figure*}

\begin{table}[t]
  \begin{center}
    \begin{tabular}{l| c}
      \textbf{Inference Pipeline} & \textbf{Geomean Speedup}\\\hline \hline
      i) Greedy Decoding (baseline) & 1.00$\times$\\ \hline
     ii) SD with Qwen2.5-Coder 0.5B draft & 2.64$\times$\\ \hline
    iii) SD with MXFP4 draft & 2.03$\times$\\ \hline
    iv) Multi-level SD & \textbf{2.72$\times$}\\ \hline
    \end{tabular}
  \end{center}
\caption{Speculative decoding speedups on code generation with Qwen2.5-Coder 7B BF16 model as baseline. We show with bold text the best performing strategy.}
\label{tab:qwen}
\end{table}

For the evaluation on code generation we benchmarked the following generation strategies: (i) BF16 Greedy decoding with Qwen2.5-Coder 7B (i.e.\ our baseline), (ii) Speculative decoding using the Qwen2.5-Coder 0.5B model as draft, (iii) Speculative decoding using an MXFP4-direct-cast model of the 7B target model as draft, and (iv) Multi-level speculative decoding with the first-level draft being an MXFP4-direct-cast model of the 7B target model and the Qwen2.5-Coder 0.5B model as second-level draft, also quantized to MXFP4. This setup with Qwen2.5-Coder models is essentially different from the one used for QA with Llama2 models: The foundational family of Qwen2.5-Coder models includes a small 0.5B model which has shown to produce exceptional results for its size~\cite{hui2024qwen2}. Therefore it may serve as a perfect draft in a SD pipeline with a single-level of drafting.

Table~\ref{tab:qwen} illustrates a summary of the experimental results. First we observe that that SD with the 0.5B draft yields a geomean speedup of 2.64$\times$ over the greedy decoding of the 7B models. The SD with the MXFP4-direct-cast model of the 7B target model as draft yields comparable speedup of 2.03$\times$. To further get some insights on these result, we illustrate in Figure~\ref{fig:accept_qwen} the draft acceptance ratios for these two SD setups over the 100 input prompts. We see that the smaller 0.5B model draft yields very high acceptance ratios for the draft tokens ($\approx$ 62\%) while being effectively 14$\times$ smaller than the 7B BF16 target and 3.7$\times$ smaller than the MXFP4-7B draft (the latter achieving $\approx 91\%$ acceptance ratios). Looking at the SD speedup projections in Figure~\ref{fig:rationale} we see that such a small draft (about 1 order of magnitude faster than the base) with acceptance ratios $\approx$ 60\% should yield speedups in the regime of 2.5$\times$ -- 3$\times$ (see light green surface band). Therefore it is expected for such a draft to outperform the MXFP4-direct-cast model of the 7B target model (see discussion in Section~\ref{subsec:limitations}). We re-iterate here that having such a small, high-quality model in a foundational family of LLMs that can be readily used as draft is not commonplace. For example, the smallest model in the foundational Code-Llama family of LLM models is 7B parameters~\cite{roziere2023code}. Nevertheless, by using this 0.5B model in a multi-level SD setup whereas the 0.5B model itself is direct-cast-quantized to MXFP4 and serves as the last-level draft, with the 7B MXFP4-quantized model being the first-level draft we achieve the best overall geomean speedup (2.72$\times$ over the 7B BF16 baseline with greedy decoding).

\subsection{MXPF4 GEMM performance and Ablation study}
\label{subsec:mxfp4_perf}
\begin{figure}[t]
\centering
\includegraphics[width= \columnwidth]{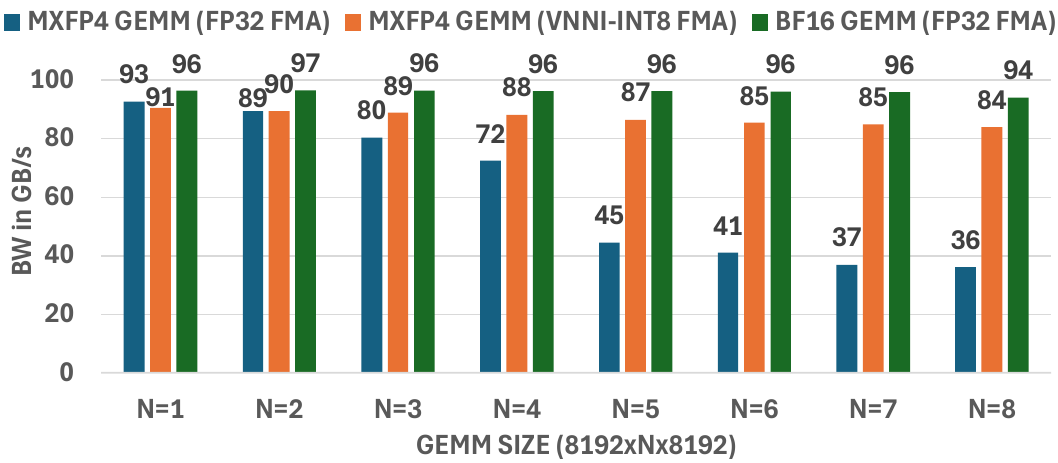}
\caption{Achieved bandwidth in GB/s for GEMM kernels with varying $N$ ($M=8192$, $K=8192$).}
\label{fig:gemm_kernel}
\end{figure}

\begin{table}[t]
  \begin{center}
    \begin{tabular}{c|c|c}
      \textbf{Inference Pipeline}
      &
      \multicolumn{2}{c}{\textbf{Speedup over Greedy}}\\
      & \textbf{\textit{FP32-FMA}} & \textbf{\textit{INT8-FMA}} \\
      \hline
      SD with MXFP4 draft & 1.90 & 2.03  \\
      \hline
      Multi-level SD & 2.16 & 2.72  \\
    \end{tabular}
  \end{center}
\caption{Speculative decoding speedups on code generation
over the Qwen2.5-Coder 7B BF16 model baseline. We illustrate the impact of the FMA type in the MXFP4-GEMM microkernel on overall performance.}
\label{tab:ablation}
\end{table}

In this section we show the performance of the MXFP4 GEMM microkernel described in Section~\ref{subsubsec:mxfp4_gemm}. In particular, since we focus on memory-bound scenarios, our objective is for our GEMM kernels to achieve the maximum bandwidth. We are not only interested in case of matrix-vector multiplication (i.e.\ GEMM with $N=1$), but also in cases with small $N$ values (e.g.\ $N \leq 8$) since these shapes are used during the draft-verification, where $N$ drafts are verified at once. 

\begin{figure}[t]
\centering
\includegraphics[width=\columnwidth]{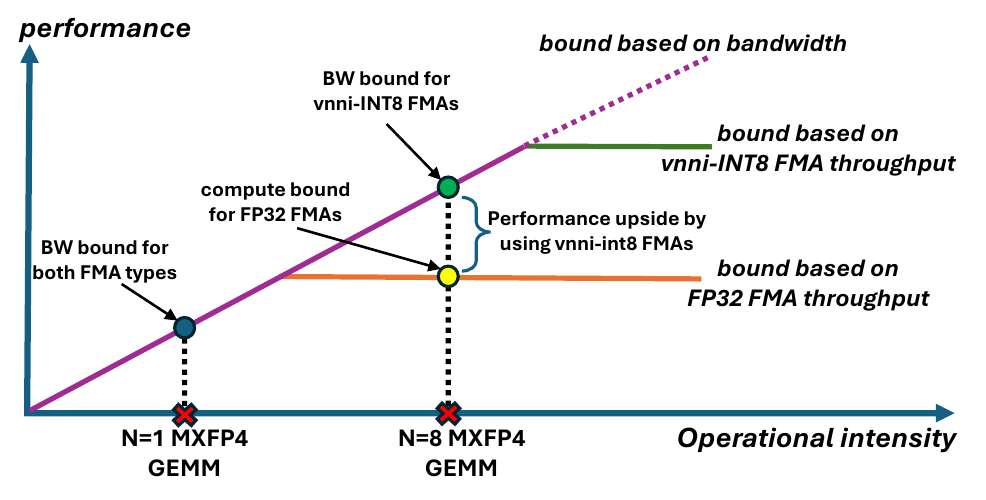}
\caption{Roofline illustrating the impact of FMA type in the MXFP4 GEMM perfomance as the N increases from 1 to 8.}
\label{fig:roofline}
\end{figure}

Figure~\ref{fig:gemm_kernel} shows the achieved bandwidth in GB/s for GEMM kernels with varying $N$ ($M=8192$, $K=8192$). Blue bars correspond to MXFP4-weight GEMMs with FP32 FMAs, orange bars depict MXFP4-weight GEMMs with vnni-INT8 FMAs (i.e.\ kernel in Figure~\ref{fig:microkernel}), and green bars show BF16-weight GEMMs with FP32 FMAs. First we see that the BF16 GEMM kernels achieve close to peak of the platform's peak BW (92\% of peak) for all $N$ values. These data dictate that the verification of the $N$ first-level draft tokens can happen by the BF16 target model in full-speed (i.e.\ at the same cost as generating 1 new token in a greedy, auto-regressive fashion). With respect to the MXFP4-weight GEMMs we see that their performance is greatly impacted by the FMA type especially for larger values of $N$. For example, for $N=8$ the MXFP4 GEMM with vnni-int8 FMAs achieves bandwidth of 84 GB/s (81\% of peak) and is 2.3$\times$ faster than the MXFP4 GEMM with FP32 FMAs. This behavior is a result of the increased operational intensity for larger $N$ values that the GEMM exhibits (see roofline model in Figure~\ref{fig:roofline}). In the case of low FMA throughput (as is the case on this platform with FP32 FMAs) and $N=8$, the kernel starts being compute bound since the computational roof peak is quite low (orange horizontal line in Figure~\ref{fig:roofline}). By employing the vnni-int8 FMAs with the necessary microkernel modifications described in Section~\ref{subsubsec:mxfp4_gemm}, we essentially raise the computational roof line by 4$\times$ (green horizontal line in Figure~\ref{fig:roofline}) and the MXFP4 GEMM kernel with $N=8$ remains in the memory-bound region, resulting in increased performance.

Table~\ref{tab:ablation} shows the results of the ablation study where we perform SD with MXFP4 draft and multi-level SD on the code-generation setup of Section~\ref{subsec:code_exp} on two different cases: using the MXFP4 GEMM kernels with FP32 FMAs and vnni-INT8 FMAs. We show that for the single level SD we observe a performance upside of 7\% (see obtained speedup of 1.9 vs 2.03). However, in the multi-level SD where the first-level MXFP4 draft model subsumes the validation of the last-level draft-generated tokens, the performance upside is 26\% (see obtained speedup of 2.16 vs 2.72). This is explained by the fact that in the last case the MXFP4 intermediate draft is using MXFP4-weight GEMMs with larger $N$ values to validate the last-level draft tokens, and these cases are more performance-sensitive to the FMA type. These results highlight that high-performance, low-precision GEMM kernels are paramount in achieving best possible performance in ML-SpecQD. 

\section{Conclusions And Future Work}
\label{sec:conclusions}
In this work we propose using MXFP4 models as draft models in a plug-and-play manner: the MXFP4 Weight-Only-Quantization merely direct-casts the BF16 target model weights to MXFP4. Our plug-and-play solution gives speedups up to 2$\times$ over the BF16 baseline. Then we propose ML-SpecQD: Multi-Level Speculative Decoding with Quantized Drafts where the MXFP4 draft token generation itself can be accelerated via SD by using another smaller draft. Combining Multi-Level SD with MXFP4 Quantized Drafts we outperform SOTA SD, yielding speedups up to 2.72$\times$ over the BF16 baseline. In simpler terms, we show that MXFP4 quantized drafts significantly lower the barrier for SD. Furthermore, combining quantized drafts with tailored, smaller pre-trained drafts – when available – yields further acceleration.

We have identified a few directions as future work. First, we plan to investigate the benefits of using recent developments in ultra-low bit LLM quantization with 1 and 2 bit models as drafts in ML-SpecQD. In the same spirit, we will investigate the impact of using simpler models as last-level drafts, e.g.\ small bigram/trigram statistical models instead of traditional LLM models. Additionally, we can embrace within ML-SpecQD the usage of joint weight-activation quantization, including KV-cache quantization to accommodate the acceleration of the use-cases with longer context~\cite{lin2024duquant,lin2024qserve,tiwari2025quantspec}. Next, our framework is platform-agnostic, and we will look into the performance upside with ML-SpecQD when considering GPUs as the underlying inference platform. Finally, implementing ML-SpecQD in production LLM serving frameworks like vLLM~\cite{kwon2023efficient} with advanced features like continuous batching is yet another exciting future direction.

\bibliographystyle{unsrt}
\bibliography{references}

\scriptsize
\noindent
\newline Optimization Notice: Software and workloads used in
performance tests may have been optimized for performance only on
Intel microprocessors.  Performance tests, such as SYSmark and
MobileMark, are measured using specific computer systems,
components, software, operations and functions.  Any change to any
of those factors may cause the results to vary.  You should
consult other information and performance tests to assist you in
fully evaluating your contemplated purchases, including the
performance of that product when combined with other products.
For more information go to http://www.intel.com/performance.

\noindent Intel, Xeon, and Intel Xeon Phi are trademarks of Intel Corporation in the U.S. and/or other countries.

\normalsize
\end{document}